\documentclass[11pt,a4paper]{article}
\usepackage[hyperref]{acl2019}
\usepackage{times}
\usepackage{latexsym}
\usepackage{url}
\usepackage{multirow}
\usepackage{adjustbox}
\usepackage[ruled]{algorithm2e} %alog
\usepackage{amssymb}% http://ctan.org/pkg/amssymb
\usepackage{pifont}% http://ctan.org/pkg/pifont
\usepackage{url}
\usepackage{mwe}
\usepackage{amsmath}
\usepackage{amssymb}
\usepackage{wasysym}
\usepackage{bbm}
\usepackage{todonotes} % insert [disable] to disable all notes
\usepackage{graphicx}
\usepackage{caption}
\usepackage[caption=false]{subfig}
\usepackage{comment}
\usepackage{multirow}
\usepackage{multicol}
\usepackage{array}

\aclfinalcopy % Uncomment this line for the final submission
\DeclareGraphicsExtensions{.pdf,.png,.jpg}

\newcommand{\lmt}{\textit{LMT}}
\newcommand{\smt}{\textit{SMT}}
\newcommand{\mmt}{\textit{M-MTL}}
\newcommand{\nmodel}{HNN}
\newcommand{\wsc}{WSC}

\title{A Hybrid Neural Network Model for Commonsense Reasoning}

\author{Pengcheng He$^1$, Xiaodong Liu$^2$, Weizhu Chen$^1$, Jianfeng Gao$^2$\\
  $^1$ Microsoft Dynamics 365 AI~
  $^2$ Microsoft Research~~~~~~~~\\
  {\tt \{penhe,xiaodl,wzchen,jfgao\}@microsoft.com}}
\begin{document}
\maketitle
\begin{abstract}
This paper proposes a hybrid neural network (HNN) model for commonsense reasoning. 
An HNN consists of two component models, a masked language model and a semantic similarity model, which share a BERT-based contextual encoder but use different model-specific input and output layers. 
HNN obtains new state-of-the-art results on three classic commonsense reasoning tasks, pushing the WNLI benchmark to 89\%, the Winograd Schema Challenge (WSC) benchmark to 75.1\%, and the PDP60 benchmark to 90.0\%.
An ablation study shows that language models and semantic similarity models are complementary approaches to commonsense reasoning, and HNN effectively combines the strengths of both.
The code and pre-trained models will be publicly available at \url{https://github.com/namisan/mt-dnn}.
\end{abstract}
%\footnote{There are 285 problems in the WSC dataset. For a fair comparison, we chose WSC273 which is a subset of WSC including 273 problems and is used by many baselines.}
%\xiaodl{WSC includes 285 samples while WSC273 includes 273 samples. I replaced WSC273 with WSC in the paper.}
\section{Introduction}
\label{sec:introduction}
%While using language model can solve 70\% of the WSC problems, we propose a new way to solve similar problem via semantic matching and learn the model jointly via MultiTask learning.

Commonsense reasoning is fundamental to natural language understanding (NLU). As shown in the examples in Table \ref{tab:example}, in order to infer what the pronoun ``they'' refers to in the first two statements, one has to leverage the commonsense knowledge that ``demonstrators usually cause violence and city councilmen usually fear violence.'' Similarly, it is obvious to humans what the pronoun ``it'' refers to in the third and fourth statements due to the commonsense knowledge that ``An object cannot fit in a container because either the object (trophy) is too big or the container (suitcase) is too small.''

\begin{table}[!ht]
\begin{enumerate} %[leftmargin=*]

\item \textit{The city councilmen refused the demonstrators a permit because \textbf{they} feared violence.} 
Who feared violence? \\
A. \textbf{The city councilmen}\hspace{0.02\textwidth} B. The demonstrators

\item \textit{The city councilmen refused the demonstrators a permit because \textbf{they} advocated violence.} 
Who advocated violence? \\
A. The city councilmen\hspace{0.02\textwidth} B. \textbf{The demonstrators}

\item \textit{The trophy doesn't fit in the brown suitcase because \textbf{it} is too big.}
What is too big? \\
A. \textbf{The trophy}\hspace{0.02\textwidth} B. The suitcase

\item \textit{The trophy doesn't fit in the brown suitcase because \textbf{it} is too small.}
What is too small? \\
A. The trophy\hspace{0.02\textwidth} B. \textbf{The suitcase}

\end{enumerate}
\caption{Examples from Winograd Schema Challenge (WSC). The task is to identify the reference of the pronoun in bold.}
\label{tab:example}
\end{table}

In this paper, we study two classic commonsense reasoning tasks: the Winograd Schema Challenge (WSC) and Pronoun Disambiguation Problem (PDP) \cite{levesque2011winograd,davis2015commonsense}. 
Both tasks are formulated as an anaphora resolution problem, which is a form of co-reference resolution, where a machine (AI agent) must identify the antecedent of an ambiguous pronoun in a statement. 
WSC and PDP differ from other co-reference resolution tasks \cite{soon2001machine,ng2002improving,peng2016event} in that commonsense knowledge, which cannot be explicitly decoded from the given text, is needed to solve the problem, as illustrated in the examples in Table \ref{tab:example}.

Comparing with other commonsense reasoning tasks, such as 
COPA \cite{roemmele2011choice}, 
Story Cloze Test \cite{mostafazadeh-EtAl:2016:N16-1},
Event2Mind \cite{rashkin2018event2mind},
SWAG \cite{zellers2018swag}, 
ReCoRD \cite{zhang2018record}, and so on, 
WSC and PDP better approximate real human reasoning, can be easily solved by native English-speaker \cite{levesque2011winograd}, and yet are challenging for machines. For example, the WNLI task, which is derived from WSC, is considered the most challenging NLU task in the General Language Understanding Evaluation (GLUE) benchmark \cite{wang2018glue}. Most machine learning models can hardly outperform the naive baseline of majority voting (scored at 65.1) \footnote{See the GLUE leaderboard at \url{https://gluebenchmark.com/leaderboard}}, including BERT \cite{devlin2018bert} and Distilled MT-DNN \cite{liu2019mt-dnn-kd}.

While traditional methods of commonsense reasoning rely heavily on human-crafted features and knowledge bases \cite{D12-1071,Sharma:2015:TAW:2832415.2832433,KR147958,SSS1510295,liu2016combing}, we explore in this study machine learning approaches using deep neural networks (DNN). Our method is inspired by two categories of  DNN models proposed recently. 

The first are neural language models trained on large amounts of text data. \citet{trinh2018simple} proposed to use a neural language model trained on raw text from books and news to calculate the probabilities of the natural language sentences which are constructed from a statement by replacing the to-be-resolved pronoun in the statement with each of its candidate references (antecedent), and then pick the candidate with the highest probability as the answer. \citet{kocijan2019surprisingly} showed that a significant improvement can be achieved by fine-tuning a pre-trained masked language model (BERT in their case) on a small amount of WSC labeled data.

The second category of models are semantic similarity models. \citet{wang-etal-2019-unsupervised} formulated WSC and PDP as a semantic matching problem, and proposed to use two variations of the Deep Structured Similarity Model (DSSM) \cite{huang2013dssm} to compute the semantic similarity score between each candidate antecedent and the pronoun by (1) mapping the candidate and the pronoun and their context into two vectors, respectively, in a hidden space using deep neural networks, and (2) computing cosine similarity between the two vectors.  The candidate with the highest score is selected as the result.

The two categories of models use different inductive biases when predicting outputs given inputs, and thus capture different views of the data. While language models measure the semantic coherence and wholeness of a statement where the pronoun to be resolved is replaced with its candidate antecedent, DSSMs measure the semantic relatedness of the pronoun and its candidate in their context. 

Therefore, inspired by multi-task learning \cite{caruana1997multitask,liu2015mtl,liu2019multi}, we propose a hybrid neural network (HNN) model that combines the strengths of both neural language models and a semantic similarity model. 
As shown in Figure \ref{fig:hybrid-model}, HNN consists of two component models, a masked language model and a deep semantic similarity model. The two component models share the same text encoder (BERT), but use different model-specific input and output layers. The final output score is the combination of the two model scores. 
The architecture of HNN bears a strong resemblance to that of Multi-Task Deep Neural Network (MT-DNN) \cite{liu2019multi}, which consists of a BERT-based text encoder that is shared across all tasks (models) and a set of task (model) specific output layers.  
Following \cite{liu2019multi,kocijan2019surprisingly}, the training procedure of HNN consists of two steps: (1) pretraining the text encoder on raw text \footnote {In this study we use the pre-trained BERT large models released by the authors.}, and (2) multi-task learning of HNN on WSCR which is the most popular WSC dataset, as suggested by \citet{kocijan2019surprisingly}.

HNN obtains new state-of-the-art results with significant improvements on three classic commonsense reasoning tasks, pushing the WNLI benchmark in GLUE to 89\%, 
the {\wsc} benchmark \footnote{\url{https://cs.nyu.edu/faculty/davise/papers/WinogradSchemas/WS.html}} 
\cite{levesque2011winograd} to 75.1\%, 
and the PDP-60 benchmark \footnote{\url{https://cs.nyu.edu/faculty/davise/papers/WinogradSchemas/PDPChallenge2016.xml}} 
to 90.0\%.
We also conduct an ablation study which shows that language models and semantic similarity models provide complementary approaches to commonsense reasoning, and HNN effectively combines the strengths of both.

\section{The Proposed HNN Model}
\label{sec:method}
\begin{figure*}
	\centering
	\vspace{-1mm}
	% \adjustbox{trim={0.0\width} {0.71\height} {0.\width} {0.01\height},clip}
    {
 	\includegraphics[width=0.75\textwidth]{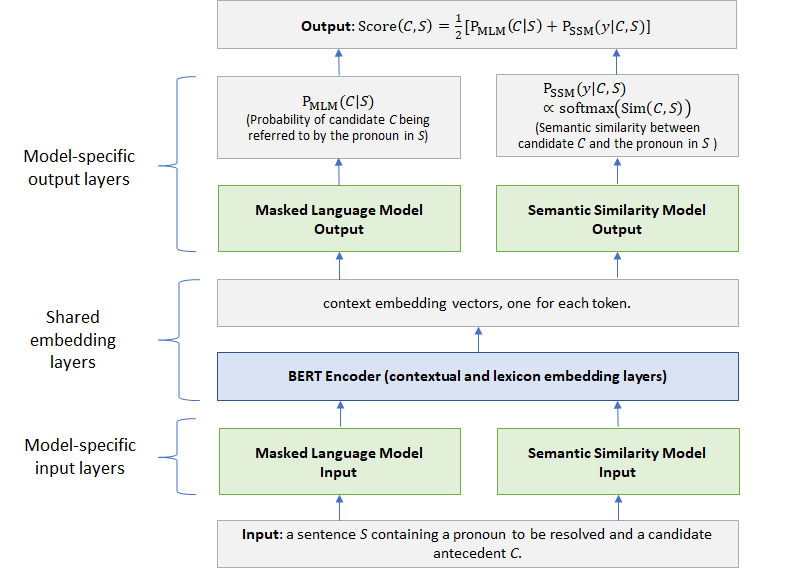}

    }
	%\vspace{-2mm}
	\caption{Architecture of the hybrid model for commonsense reasoning. The model consists of two component models, a masked language model (MLM) and a semantic similarity model (SSM). The input includes the sentence $S$, which contains a pronoun to be resolve, and a candidate antecedent $C$. The two component models share the BERT-based contextual encoder, but use different model-specific input and output layers. The final output score is the combination of the two component model scores.}
    %\vspace{-4mm}
	\label{fig:hybrid-model}
\end{figure*}

The architecture of the proposed hybrid model is shown in Figure \ref{fig:hybrid-model}. The input includes a sentence $S$, which contains the pronoun to be resolved, and a candidate antecedent $C$. The two component models, masked language model (MLM) and semantic similarity model (SSM), share the BERT-based contextual encoder, but use different model-specific input and output layers. The final output score, which indicates whether $C$ is the correct candidate of the pronoun in $S$, is the combination of the two component model scores.

\subsection{Masked Language Model (MLM)}
\label{sec:mlm}

This component model follows \citet{kocijan2019surprisingly}.
In the input layer, a masked sentence is constructed using $S$ by replacing the to-be-resolved pronoun in $S$ with a sequence of $N$ \texttt{[MASK]} tokens, where $N$ is the number of tokens in candidate $C$.

In the output layer, the likelihood of $C$ being referred to by the pronoun in $S$ is scored using the BERT-based masked language model $P_{mlm}(C|S)$. If $C=\{c_1...c_N\}$ consists of multiple tokens, $\log P_{mlm} (C|S)$ is computed as the average of log-probabilities of each composing token:
\begin{equation}
P_{mlm}(C|S)= \exp{ \left ( \frac{1}{N} \sum_{k=1...N} \log P_{mlm}(c_k|S) \right )}.
\label{eqn:mlm}
\end{equation}

\subsection{Semantic Similarity Model (SSM)}
\label{sec:ssm}

In the input layer, we treat sentence $S$ and candidate $C$ as a pair $(S,C)$ that is packed together as a word sequence, where we add the \texttt{[CLS]} token as the first token and the \texttt{[SEP]} token between $S$ and $C$.

After applying the shared embedding layers, we obtain the semantic representations of $S$ and $C$, denoted as $\mathbf{s}\in \mathbb{R}^d$ and $\mathbf{c} \in \mathbb{R}^d$, respectively. We use the contextual embedding of \texttt{[CLS]} as $\mathbf{s}$. Suppose $C$ consists of $N$ tokens, whose contextual embeddings are $\mathbf{h}_1,...,\mathbf{h}_N$, respectively. The semantic representation of the candidate $C$, $\mathbf{c}$, is computed via attention as follows:
\begin{equation}
\alpha_k = \text{softmax} (\frac{\mathbf{s}^\top \mathbf{W}_1 \mathbf{h}_k}{\sqrt{d}}),
\label{eqn:attn1}
\end{equation}
\begin{equation}
\mathbf{c} = \sum_{k=1...N} {\alpha_k \cdot \mathbf{h}_k}.
\label{eqn:attn2}
\end{equation}
where $\mathbf{W}_1$ is a learnable parameter matrix, and $\alpha$ is the attention score. 

We use the contextual embedding of the first token of the pronoun in $S$ as the semantic representation of the pronoun, denoted as $\mathbf{p} \in \mathbb{R}^d$. 
In the output layer, the semantic similarity between the pronoun and the context is computed using a bilinear model:
\begin{equation}
\text{Sim}(C,S) = \mathbf{p}^\top \mathbf{W}_2 \mathbf{c},
\label{eqn:ssm-m}
\end{equation}
where $\mathbf{W}_2$ is a learnable parameter matrix.
Then, SSM predicts whether $C$ is a correct candidate (i.e., $(C,S)$ is a positive pair, labeled as $y=1$) using the logistic function:
\begin{equation}
% P_{ssm}(y=1|C,S) = \text{softmax} (\text{Sim}(C,S)).
P_{ssm}(y=1|C,S) = \frac{1}{1 + \exp{(-\text{Sim}(C,S))}}.
\label{eqn:ssm}
\end{equation}

The final output score of pair $(S,C)$ is a linear combination of the MLM score of Eqn. \ref{eqn:mlm} and the SSM score of Eqn. \ref{eqn:ssm}:
\begin{equation}
\text{Score}(C,S) =\frac{1}{2} [P_{mlm}(C|S) + P_{ssm}(y=1|C,S)].
\label{eqn:score}
\end{equation}
%\xiaodl{We can take mean to convert it in the range (0, 1).}
\subsection{The Training Procedure}

We train our model of Figure \ref{fig:hybrid-model} on the WSCR dataset, which consists of 1886 sentences, each being paired with a positive candidate antecedent and a negative candidate. 

The shared BERT encoder is initialized using the published BERT uncased large model \cite{devlin2018bert}. We then finetune the model on the WSCR dataset by optimizing the combined objectives:
\begin{equation}
\mathcal{L}_{mlm} + \mathcal{L}_{ssm} + \mathcal{L}_{rank},
\label{eqn:obj}
\end{equation}
where $\mathcal{L}_{mlm}$ is the negative log-likelihood based on the masked language model of Eqn.~\ref{eqn:mlm}, and $\mathcal{L}_{ssm}$ is the cross-entropy loss based on semantic similarity model of Eqn.~\ref{eqn:ssm}. 

$\mathcal{L}_{rank}$ is the pair-wise rank loss. 
Consider a sentence $S$ which contains a pronoun to be resolved, and two candidates $C^+$ and $C^-$, where $C^+$ is correct and $C^-$ is not. We want to maximize $\Delta = \text{Score}(S,C^+) - \text{Score}(S,C^-)$, where $\text{Score(.)}$ is defined by Eqn. \ref{eqn:score}. We achieve this via optimizing a smoothed rank loss:
\begin{equation}
\mathcal{L}_{rank} = \log (1 + \exp{(-\gamma (\Delta + \beta))}),
\label{eqn:maxm}
\end{equation}
where $\gamma \in [1, 10]$ is the smoothing factor and $\beta \in [0, 1]$ the margin hyperparameter. In our experiments, the default setting is $\gamma = 10$, and $\beta = 0.6$.

\section{Experiments}
\label{sec:exp}
We evaluate the proposed {\nmodel} on three commonsense benchmarks: {\wsc} \cite{winograd2012}, PDP60\footnote{\url{https://cs.nyu.edu/faculty/davise/papers/WinogradSchemas/PDPChallenge2016.xml}} and WNLI. 
WNLI is derived from WSC, and is considered the most challenging NLU task in the GLUE benchmark \cite{wang2018glue}.
%which is constructed from {\wsc}  and is deemed as the most challenging task on GLUE benchmarks \cite{wang2018glue}.  

\subsection{Datasets}
\label{subsec:data}
\begin{table}[htb!]
	\begin{center}
		\begin{tabular}{l|c|c|c}
			\hline \bf Corpus & \#Train & \#Dev & \#Test\\ \hline \hline
            WNLI& - & 634 + 71& 146  \\ \hline
            PDP60 &-& -& 60  \\ \hline
            {\wsc} &-& -& 285  \\ \hline
            WSCR &1322&564 &-   \\ \hline
		\end{tabular}
	\end{center}
%	\lgspace
	\caption{Summary of the three benchmark datasets: {\wsc}, PDP60 and WNLI, and the additional dataset WSCR. Note that we only use WSCR for training. For WNLI, we merge its official training set containing 634 instances and dev set containing 71 instances as its final dev set.
	}
	\label{tab:datasets}
\end{table}
Table~\ref{tab:datasets} summarizes the datasets which are used in our experiments. 
% As we observe from Table~\ref{tab:datasets}, 
Since the {\wsc} and PDP60 datasets do not contain any training instances, following \cite{kocijan2019surprisingly}, we adopt the WSCR dataset \cite{rahman2012resolving} for model training and selection. 
%The WSCR dataset contains 1886 instances with each instance following the same structure as WSC. 
WSCR contains 1886 instances (1322 for training and the rest as dev set). 
Each instance is presented using the same structure as that in WSC. 
%1322 instances in the training set and 564 instances in the dev set. 
%We split it to training and development set, which includes 1322 and 564 instances for training and development, respectively.
%The data set is important for this kind of tasks, following \cite{kocijan2019surprisingly}, we use the WSCR data set \cite{rahman2012resolving} for training. It contains 1886 sentences each with two antecedent candidates. As the data collection procedure is very close to WSC273, when fine-tuning on top of BERT \cite{devlin2018bert}, it boosts the performance of WSC273, as in \cite{kocijan2019surprisingly}. WSCR is separated into 1322 training data and 564 test data. We use the same split to compare the performance of our model with others. 

For the WNLI instances, we convert them to the format of {\wsc} as illustrated in Table~\ref{tab:convert}: we first detect pronouns in the premise using spaCy\footnote{\url{https://spacy.io}}; then given the detected pronoun, we search its left of the premise in hypothesis to find the longest common substring (LCS) ignoring character case. Similarly, we search its right part to the LCS; by comparing the indexes of the extracted LSCs, we extract the candidate. A detailed example of the conversion process is provided in Table~\ref{tab:convert}.
%we find the longest common substring between the premise and its hypothesis, and last by comparing it with the hypothesis
%\WZ{we need to make it more clear}, we extract the candidate. 
% Table~\ref{tab:convert} shows an example how we convert three WNLI instances into a WSC instance.
%\xiaodl{I made some change here based on weizhu comments.}
%There are 46 unique premises with 51 premise-pronoun pairs in WNLI test data. Instead of extracting all 51 pronouns in premise as candidates, we use the candidates in hypothesis of the same premise-pronoun pair as the candidates. \xiaodl{pengcheng: what does it mean? how many pairs are contructed?}
%We use spaCy\footnote{\url{https://spacy.io}} to detect the pronoun in the premise sentences of WNLI task and extract candidates in the hypothesis by longest string matching. There are 46 premise with 51 premise-pronoun pairs in WNLI test data. Instead of extracting all nouns in premise as candidates, we use the candidates in hypothesis of the same premise-pronoun pair as the candidates.
\begin{table}[!ht]
\begin{enumerate} %[leftmargin=*]

\item \textbf{Premise:} The cookstove was warming the kitchen, and \textit{\textcolor{brown}{the lamplight made} \textbf{it} \textcolor{violet}{seem even warmer.}} \\
\textbf{Hypothesis:} \textit{\textcolor{brown}{The lamplight made} \textbf{the cookstove} \textcolor{violet}{seem even warmer.}} \\
\textbf{Index of LCS in the hypothesis:} left[0, 2], right[5, 7] \\
\textbf{Candidate:} [3, 4] (the cookstove)

\item \textbf{Premise:} The cookstove was warming the kitchen, and \textit{\textcolor{brown}{the lamplight made} \textbf{it} \textcolor{violet}{seem even warmer.}} \\
\textbf{Hypothesis:} \textit{\textcolor{brown}{The lamplight made} \textbf{the kitchen} \textcolor{violet}{seem even warmer.}} \\
\textbf{Index of LCS in the hypothesis:} left[0, 2], right[5, 7] \\
\textbf{Candidate:} [3, 4] (the kitchen)

\item \textbf{Premise:} The cookstove was warming the kitchen, and \textit{\textcolor{brown}{the lamplight made} \textbf{it} \textcolor{violet}{seem even warmer.}} \\
\textbf{Hypothesis:} \textit{\textcolor{brown}{The lamplight made} \textbf{the lamplight} \textcolor{brown}{seem even warmer.}}\\
\textbf{Index of LCS in the hypothesis:} left[0, 2], right[5, 7] \\
\textbf{Candidate:} [3, 4] (the lamplight)

\item \textbf{\textcolor{blue}{Converted:}} The cookstove was warming the kitchen, and \textit{the lamplight made \textbf{it} seem even warmer.} 
 \\
A. the cookstove\hspace{0.01\textwidth} B. \textbf{the kitchen} \hspace{0.01\textwidth} C. the lamplight

\end{enumerate}
\caption{Examples of transforming WNLI to {\wsc} format. Note that the text highlighted by \textcolor{brown}{brown} is the longest common substring from the left part of pronoun \textit{it}, and the text highlighted by \textcolor{violet}{violet} is the longest common substring from its right.}
\label{tab:convert}
\end{table}

\subsection{Implementation Detail}
\label{exp:imp}
Our implementation of {\nmodel} is based on the PyTorch implementation of BERT\footnote{\url{https://github.com/huggingface/pytorch-pretrained-BERT}}. All the models are trained with hyper-parameters depicted as follows unless stated otherwise.
The shared layer is initialized by the BERT uncased large model. Adam \cite{kingma2014adam} is used as our optimizer with a learning rate of 1e-5 and a batch size of 32 or 16. The learning rate is linearly decayed during training with 100 warm up steps. We select models based on the dev set by greedily searching epochs between 8 and 10. The trainable parameters, e.g., $\mathbf{W_1}$ and $\mathbf{W_2}$, are initialized by a truncated normal distribution with a mean of $0$ and a standard deviation of $0.01$.
The margin hyperparameter, $\beta$ in Eqn.~\ref{eqn:maxm}, is set to 0.6 for MLM and 0.5 for SSM, and $\gamma$ is set to 10 for all tasks.  
We also apply SWA \cite{izmailov2018averaging} to improve the generalization of models. All the texts are tokenized using WordPieces, and are chopped to spans containing 512 tokens at most.
%Following \cite{kocijan2019surprisingly}, our models are built on top of BERT \cite{devlin2018bert} and use BERT uncased large model as the initialization. We train our model with WSCR with learning rate $1e-5$, linear learning rate decay, warm up steps of 100, batch size $\{16,32\}$, for $\{8,10\}$ epochs. Following \cite{devlin2018bert}, we use truncated norm with mean $0$, std $0.01$ to initialize $\mathbf{W_1}$ and $\mathbf{W_2}$. As the training data set is small, we apply SWA \cite{izmailov2018averaging} to improve generalization of the model. All other settings are the same as \cite{devlin2018bert}. 

%We use WSCR test data for model validation and selection. We also merged WNLI training data and dev data to scan the threshold of binary classification for WNLI task.

\subsection{Results}
\label{exp:results}

We compare our {\nmodel} with a list of state-of-the-art models in the literature, including BERT \cite{bert2018}, GPT-2 \cite{radford2019language} and DSSM \cite{wang-etal-2019-unsupervised}. The brief description of each baseline is introduced as follows.
\begin{enumerate}
    \item BERT\textsubscript{LARGE-LM} \cite{bert2018}: This is the large BERT model, and we use MLM to predict a score for each candidate following Eq~\ref{eqn:mlm}.
    \item GPT-2 \cite{radford2019language}: During prediction, We first replace the pronoun in a given sentence with its candidates one by one.  We use the GPT-2 model to compute a score for each new sentence after the replacement, and select the candidate with the highest score as the final prediction. 
   \item BERT\textsubscript{Wiki-WSCR} and BERT\textsubscript{WSCR} \cite{kocijan2019surprisingly}: These two models use the same approach as BERT\textsubscript{LARGE-LM}, but are trained with different additional training data. For example, BERT\textsubscript{Wiki-WSCR} is firstly fine-tuned on the constructed Wikipedia data and then on WSCR. 
   BERT\textsubscript{WSCR} is directly fine-tuned on WSCR.
  \item{DSSM} \cite{wang-etal-2019-unsupervised}: It is the unsupervised semantic matching model trained on the dataset generated with heuristic rules.
  \item {\nmodel}: It is the proposed hybrid neural network model.
\end{enumerate}
\begin{table*}[htb!]
	\begin{center}
		\begin{tabular}{l |c c c} \hline
			 &WNLI & {\wsc}  & PDP60 \\ \hline \hline
			DSSM \cite{wang-etal-2019-unsupervised}    &- &63.0 & 75.0 \\  \hline \hline
            
			BERT$_{\text{LARGE-LM}}$ \cite{devlin2018bert}& 65.1 &62.0  &78.3\\ \hline
			GPT-2 \cite{radford2019language}   &- &70.7 & - \\ \hline
            BERT$_{\text{Wiki-WSCR}}$ \cite{kocijan2019surprisingly} &71.9 &72.2 & - \\ \hline	   
            BERT$_{\text{WSCR}}$ \cite{kocijan2019surprisingly} &70.5 &70.3 & - \\	    \hline \hline 
            {\nmodel} &\bf 83.6 & \bf 75.1 &  \bf 90.0\\\hline 
            {\nmodel}$_{\text{ensemble}}$ &\bf 89.0 & - &  - \\\hline
            % M-MTL$_{\text{cls}}$\\(Threshold=0.47) &\bf 81.5 & - &  -\\ 
            % M-MTL$_{\text{ranking}}$ &\bf 83.6 & \bf 75.1 &  \bf 90.0\\\hline
            % M-MTL$_{\text{cls\_ensemble}}$ &\bf 81.5 & - &  -\\ 
            % M-MTL$_{\text{ranking\_ensemble}}$ &\bf 89.0 & - &  - \\\hline
		\end{tabular}
	\end{center}
    \caption{Test results}
	\label{tab:test}
\end{table*}

\begin{figure*}[ht!]
	\centering

    {
	\includegraphics[width=1.0\textwidth]{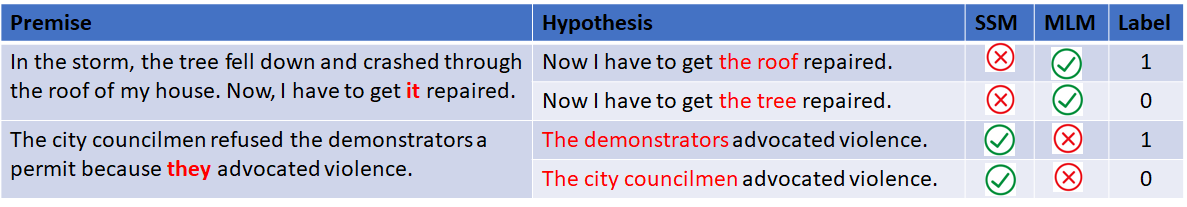}
    }
	\caption{Comparison with SSM and MLM on WNLI examples.}
	\label{fig:sample}
\end{figure*}

%We evaluate our model on three tasks, WNLI, WSC273, PDP60. In addition of the parameter setting in \ref{exp:imp}, we select $\beta=0.6$ for \lmt~and \mmt, $\beta=0.5$ for \smt, $\gamma=10$ for all tasks.  All the three models are trained for 8 epochs with batch size of 16. 

\begin{table}[htb!]
	\begin{center}
		\begin{tabular}{@{\hskip1pt}l |c c |c c } \hline
			 &WNLI &WSCR & {\wsc} & PDP60 \\ \hline \hline
            {\nmodel} &{\bf 77.1} & \textbf{85.6} &\bf 75.1 &  \bf 90.0\\\hline
            -SSM & 74.5 & 82.4& 72.6 &  86.7\\\hline
            -MLM &75.1 & 83.7& 72.3 & 88.3\\\hline
		\end{tabular}
	\end{center}
    \caption{Ablation study of the two component model in {\nmodel}. Note that WNLI and WSCR are reported on dev sets while WSC and PDP60 are reported on test sets.}
	\label{tab:mlm_ssm}
\end{table}

The main results are reported in Table~\ref{tab:test}. Compared with all the baselines, {\nmodel} obtains much better performance across three benchmarks. 
This clearly demonstrates the advantage of the {\nmodel} over existing models. 
For example, {\nmodel} outperforms the previous state-of-the-art BERT$_{\text{Wiki-WSCR}}$ model with a 11.7\% absolute improvement (83.6\% vs 71.9\%) on WNLI and a 2.8\% absolute improvement (75.1\% vs 72.2\%) on {\wsc} in terms of accuracy.
Meanwhile, it achieves a 11.7\% absolute improvement over the previous state-of-the-art BERT$_{\text{LARGE-LM}}$ model on PDP60 in accuracy. 
Note that both BERT$_{\text{Wiki-WSCR}}$ and BERT$_{\text{LARGE-LM}}$ are using language model-based approaches to solve the pronoun resolution problem. 
On the other hand, We observe that DSSM without pre-training is comparable to BERT$_{\text{LARGE-LM}}$ which is pre-trained on the large scale text corpus (63.0\% vs 62.0\% on WSC and 75.0\% vs 78.3\% on PDP60). %Furthermore, BERT$_{\text{WSCR}}$ outperforms BERT$_{\text{LARGE-LM}}$ in a large margin indicating that training on a proper dataset helps performance. 
Our results show that {\nmodel}, combining the strengths of both DSSM and BERT$_{\text{WSCR}}$, has consistently achieved new state-of-the-art results on all three tasks.
 %demonstrating its effectiveness and robustness of MTL training. 
 %It is worth noting that all the models except DSSM benefit from the pre-training on large-scale raw corpus. And this bring a consistent improvement on all the three tasks. This shows importance of contextual representations.  

To further boost the WNLI accuracy on the GLUE benchmark leaderboard, we record the model prediction at each epoch, and then produce the final prediction based on the majority voting from the last six model predictions. 
We refer to the ensemble of six models as {\nmodel}$_{\text{ensemble}}$ in Table~\ref{tab:test}. {\nmodel}$_{\text{ensemble}}$ brings a 5.4\% absolute improvement (89.0\% vs 83.6\%) on WNLI in terms of accuracy. 

%From the results in table \ref{tab:test}, we can see \smt~ does better than \lmt~ for WNLI task. We think this is because in WNLI, some of the candidates is not a directly copy and when directly replace pronoun with it, the sentence may not be grammar correct. To solve such kind of cases, we need semantic matching instead of simply language model acceptability. We will discuss about it with more details in section \ref{ablation:smt}

%With \mmt, we combine the benefit of \lmt~ and \smt~, which boost the performance of WNLI and WSC273 a lot. 

%For WNLI task, we run several other experiments with $\beta_2\in \{0,0.2\}$, $\gamma_3 \in \{5,10\}, \beta_3 \in \{0.5,0.6\}$. And produce ensemble results by simply majority voting.

\subsection{Ablation study}

%In this section, we will do some ablation study on the effect of \smt and \mmt.
%As we don't have label of WNLI test data, in this section, we will use WNLI taining data, DPRD test data, WSC273 test data and PDP60 test data for ablation study.
%It is crucial to understand the detailed design of the proposed model for inspiration of future work.
In this section, we study the importance of each component in {\nmodel} by answering following questions: 

\noindent \textbf{How important are the two component models: MLM and SSM?}

To answer this question, we first remove each component model, either SSM or MLM, and then report the performance impact of these component models. 
Table~\ref{tab:mlm_ssm} summarizes the experimental results. It is expected that the removal of either component model results in a significant performance drop. 
For example, with the removal of SSM, the performance of {\nmodel} is downgraded from 77.1\% to 74.5\% on WNLI. Similarly, with the removal of MLM, {\nmodel} only obtains 75.1\%, which amounts to a 2\% drop. All these observations clearly demonstrate that SSM and MLM are complementary to each other and the {\nmodel} model benefits from the combination of both. 

Figure~\ref{fig:sample} gives several examples showing how SSM and MLM complement each other on WNLI. We see that in the first pair of examples, MLM correctly predicts the label while SSM does not. This is due to the fact that ``the roof repaired'' appears more frequently  than ``the tree repaired'' in the text corpora used for model pre-training. 
However, in the second pair, since both ``the demonstrators'' and ``the city councilment'' could advocate violence and neither occurs significantly more often than the other,  SSM is more effective in distinguishing the difference based on their context. 
The proposed HNN, which combines the strengths of these two models, can obtain the correct results in both cases.
%and achieve the state-of-the-art results on these datasets. 

% \noindent \textbf{Does score ensemble help? \WZ{we may re-think about whether current approach to tell ensemble is convincing or not}}

% \begin{table}[htb]
% 	\begin{center}
% 		\begin{tabular}{l |c c |c c } \hline
% 			 &WNLI &WSCR & {\wsc} & PDP60 \\ \hline \hline
%             {\nmodel}+Eq~\ref{eqn:score} &{\bf 77.1} & 85.6 &\bf 75.1 &  \bf 90.0\\\hline
%             {\nmodel}+Eq~\ref{eqn:ssm} & 73.7 & \bf 86.3& 69.5 &  88.3\\\hline
%              {\nmodel}+Eq~\ref{eqn:mlm} &74.8 & 83.9& 72.3 & 88.3\\\hline
% 		\end{tabular}
% 	\end{center}
%     \caption{Ablation study of score ensemble (Eq~\ref{eqn:score}). Note that WNLI and WSCR are reported on dev sets while WSC and PDP60 are reported on test sets.}
% 	\label{tab:fn_s}
% \end{table}

% The second question is whether score ensemble helps as in Eq~\ref{eqn:score}. We first trained the {\nmodel} jointly as in Eq~\ref{eqn:obj}, and then we predict answers with only the MLM module or SSM module. As shown in Table~\ref{tab:fn_s}, by averaging score (Eq~\ref{eqn:score}), {\nmodel} always obtains a better result except the WSCR dev dataset. \xiaodl{We need to explain why it does not work on WSCR.}
% %It indicates that robustness of score ensemble. 

\noindent \textbf{Does the additional ranking loss help?}

%\JG{This is confusing! Do you refer to $L_{rank}$ as the extra smoothing loss in Eq. 7? I rewrite this part assuming that we want to justify the use of $L_{rank}$.} \xiaodl{Yes, we try to study Equation 8.}
As in Eqn.~\ref{eqn:obj}, the training objective of {\nmodel} model contains three losses. The first two are based on the two component models, respectively, and the third one, as defined in Eqn.~\ref{eqn:maxm}, is a ranking loss based on the score function in Eqn.~\ref{eqn:score}. At first glance, the ranking loss seems redundant. 
Thus, we compare two versions of HNN trained with and without the ranking loss.
%an extra smoothing loss.
%We design this for optimization purpose. We compare {\nmodel} with or without this additional smoothing loss. 
Experimental results are shown in Table~\ref{tab:ranking_loss}. We see that without the ranking loss, the performance of {\nmodel} drops on three datasets: WNLI, WSCR and WSC. On the PDP60 dataset, without the ranking loss, the model performs slightly better. However, since the test set of PDP60 includes only 60 samples, the difference is not statistically significant. 
Thus, we decide to always include the ranking loss in the training objective of HNN.
% use future work to add this smoothing step into the final loss function.  

\begin{table}[htb]
	\begin{center}
		\begin{tabular}{@{\hskip1pt}l |c c |c c } \hline
			 &WNLI &WSCR & {\wsc} & PDP60 \\ \hline \hline
            {\nmodel} &{\bf 77.1} & \bf 85.6 &\bf 75.1 &  90.0\\\hline
            {\nmodel}-Rank & 74.8 &85.1& 71.9 &  \bf 91.7\\\hline
		\end{tabular}
	\end{center}
    \caption{Ablation study of the ranking loss. Note that WNLI and WSCR are reported on dev sets while WSC and PDP60 are reported on test sets.}
	\label{tab:ranking_loss}
\end{table}

\noindent \textbf{Is the WNLI task a ranking or classification task?}

%\JG{This section is confusing and not necessary. Since we already use ranking loss for training, I don't know how you perform the comparison here. Do you drop the ranking loss term in Eqn.~\ref{eqn:obj} for classification models? If so, this section is not useful since we already show the ranking loss is needed. If we believe that the results here are useful, you need to articular how the classification models and ranking models are trained differently.}

%\xiaodl{My understanding is that the training is the same. During the prediction, we can use a threshold to decide the label (0/1) for each pair by using some threshold. Alternatively, as shown in Table 3, we can compute scores for these 3 pairs, and then pick the highest score pair as 1 and others are 0. Am I right? @pengcheng. BTW, Scott also asked this question.}

%\PC{Agree with Xiaodong.}

\begin{figure}[htb]
	\centering
	\adjustbox{trim={0.08\width} {0.0\height} {0.09\width} {0.01\height},clip}
    {
	\includegraphics[width=0.58\textwidth]{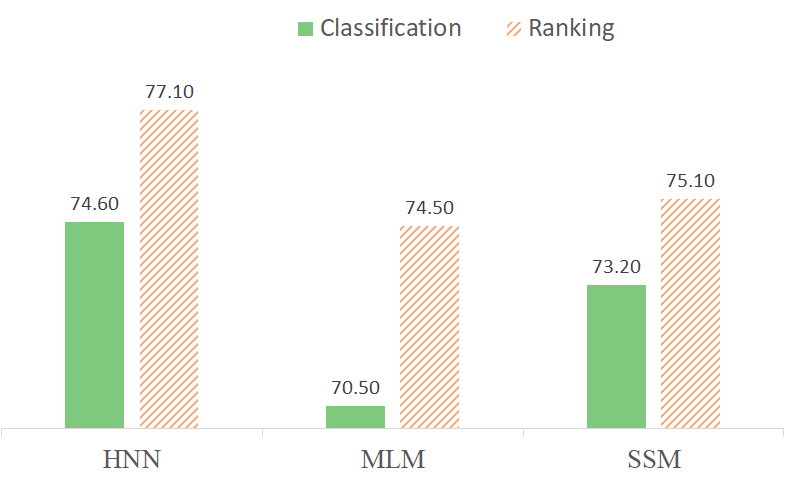}
    }
	\caption{Comparison of different task formulation on WNLI.}
    %\vspace{-4mm}
	\label{fig:rank_cls}
\end{figure}

The WNLI task can be formulated as either a ranking task or a classification task. To study the difference in problem formulation, we conduct experiments to compare the performance of a model used as a classifier or a ranker. For example, given a trained HNN, when it is used as a classifier we set a threshold to decide label (0/1) for each input. When it is used as a ranker, we simply pick the top-ranked candidate as the correct answer.
We run the comparison using all three models HNN, MLM and SSM. 
As shown in Figure~\ref{fig:rank_cls}, the ranking formulation is consistently better than the classification formulation for this task. 
For example, the difference in the HNN model is about absolute 2.5\% (74.6\% vs 77.1\%) in terms of accuracy. 

\section{Conclusion}
\label{sec:con}
We propose a hybrid neural network (HNN) model for commonsense reasoning.
HNN consists of two component models, a masked language model and a deep semantic similarity model, which share a BERT-based contextual encoder but use different model-specific input and output layers.

HNN obtains new state-of-the-art results on three classic commonsense reasoning tasks, pushing the WNLI benchmark to 89\%, the {\wsc} benchmark to 75.1\%, and the PDP60 benchmark to 90.0\%. We also justify the design of HNN via a series of ablation experiments.

In future work, we plan to extend HNN to more sophisticated reasoning tasks, especially those where large-scale language models like BERT and GPT do not perform well, as discussed in \cite{gao2019neural, niven2019probing}. 

% \input{related}

% comment it out during the submission
\section*{Acknowledgments}
We would like to thank Michael Patterson from Microsoft for his help on the paper. 

% all names here are ordered alphabetically. 

% include your own bib file like this:
%\bibliographystyle{acl}
%\bibliography{acl2018}
\bibliography{acl_snli}
\bibliographystyle{acl_natbib}
%\appendix
%
%\section{Supplemental Material}
%\label{sec:supplemental}
\end{document}